\documentclass[runningheads]{llncs}

 \usepackage[mobile]{eccv}
\usepackage{microtype}
\usepackage{eccvabbrv}
\usepackage{siunitx}

\usepackage{graphicx}
\usepackage{booktabs}
\usepackage{subcaption}

\usepackage[accsupp]{axessibility}  %

\usepackage{hyperref}

\usepackage{orcidlink}

\usepackage{array}
\newcolumntype{H}{>{\setbox0=\hbox\bgroup}c<{\egroup}@{}}
\usepackage{makecell}
\usepackage{adjustbox}
\usepackage{cases}

\usepackage{multirow}

\usepackage{pifont}

\usepackage{hyphenat}

\usepackage{placeins}

\DeclareMathOperator{\R}{\mathbb{R}}

\DeclareMathOperator{\N}{\mathbb{N}}

\makeatletter
\DeclareRobustCommand\onedot{\futurelet\@let@token\@onedot}
\def\@onedot{\ifx\@let@token.\else.\null\fi\xspace}

\def\eg{\emph{e.g}\onedot} 
\def\ie{\emph{i.e}\onedot} 
\def\cf{\emph{cf.}\xspace} 

 \def\vs{\emph{vs}\onedot}
 
\def\etal{\emph{et~al}\onedot}
\makeatother

\newcommand{\trackastra}{\mbox{\textsc{Track\-astra}}\xspace}

\newcommand{\bacteria}{\mbox{\textsc{Bacteria}}\xspace}
\newcommand{\deepcell}{\mbox{\textsc{DeepCell}}\xspace}
\newcommand{\vesicle}{\mbox{\textsc{Vesicle}}\xspace}
\newcommand{\hela}{\mbox{\textsc{Hela}}\xspace} 
\begin{document}
\sisetup{detect-weight=true,detect-inline-weight=math}

\title{\trackastra: Transformer-based cell tracking for live-cell microscopy}

\author{Benjamin Gallusser\inst{1}\orcidlink{0000-0002-7906-4714} \and
Martin Weigert\inst{1,2,3}\orcidlink{0000-0002-7780-9057}}

\institute{Institute of Bioengineering, School of Life Sciences, EPFL, Switzerland \and
Center for Scalable Data Analytics and AI (ScaDS.AI), Dresden/Leipzig, Germany \and
Faculty of Computer Science, TU Dresden, Germany\\
\email{benjamin.gallusser@epfl.ch}, \email{marweigert@gmail.com}\\
} 

\authorrunning{B.~Gallusser and M.~Weigert}

\maketitle

\begin{abstract}

Cell tracking is a ubiquitous image analysis task in live-cell microscopy. Unlike multiple object tracking (MOT) for natural images, cell tracking typically involves hundreds of similar-looking objects that can divide in each frame, making it a particularly challenging problem.
Current state-of-the-art approaches follow the tracking-by-detection paradigm, \ie first all cells are detected per frame and successively linked in a second step to form biologically consistent cell tracks.
Linking is commonly solved via discrete optimization methods, which require manual tuning of hyperparameters for each dataset and are therefore cumbersome to use in practice.
Here we propose \trackastra, a general purpose cell tracking approach that uses a simple transformer architecture to directly learn pairwise associations of cells within a temporal window from annotated data.
Importantly, unlike existing transformer-based MOT pipelines, our learning architecture also accounts for dividing objects such as cells and allows for accurate tracking even with simple greedy linking, thus making strides towards removing the requirement for a complex linking step.
The proposed architecture operates on the full spatio-temporal context of detections within a time window by avoiding the computational burden of processing dense images.
We show that our tracking approach performs on par with or better than highly tuned state-of-the-art cell tracking algorithms for various biological datasets, such as bacteria, cell cultures and fluorescent particles.
We provide code at \url{https://github.com/weigertlab/trackastra}.

\end{abstract} %
\section{Introduction}
\label{sec:intro}

The accurate tracking of cells in microscopy videos is an
 important step in many biological experiments, for example when studying the temporal dynamics of bacteria colonies, eukaryotic cell cultures, or whole-embryo development~\cite{malin-mayor2023,vanvliet2018,jaqaman2008}. 
This task not only consists of identifying the trajectories of individual cells, but as well requires to  correctly assign mother-daughter relationships of dividing cells over multiple generations, which is crucial for inference of cell lineages.
Typical live-cell videos can contain hundreds or thousands of visually similar cells that continuously divide and that exhibit various local and global movement patterns in between divisions, which contributes to the complexity of the task~(\cf~\cref{fig:data} for examples).
\begin{figure}[t]
    \centering
    \includegraphics[width=1.0\textwidth]{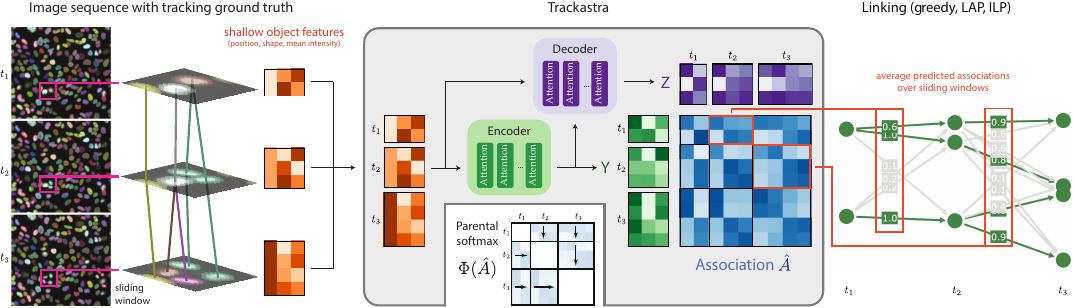}
    \caption{Overview of \trackastra. Given frame-by-frame object detections in a live-cell video, object features are extracted from a small temporal window and passed as tokens into an encoder-decoder transformer, to predict pairwise associations $\hat{A}$. We apply a \emph{parental softmax} normalization on $\hat{A}$ to guide the learning directly towards biologically plausible associations.
    Finally, we build a candidate graph by averaging the predictions $\hat{A}$ over a sliding window, and obtain a tracking solution by pruning the graph with either a greedy algorithm or discrete optimization.}
    \label{fig:method}
\end{figure}
While cell tracking has been performed manually during initial experiments of embryo development~\cite{celegans1983}, many specialized semi-automated and fully automated algorithms have been developed over the last two decades~\cite{hirsch2022tracking}. Most current state-of-the-art cell tracking methods follow the \textit{tracking-by-detection} approach,  where first cell instances are detected in each frame of a recorded video, which then in a second step are linked across frames to form biologically valid tracks.
In recent years, several robust deep-learning-based methods have emerged for the previously tedious and parameter-sensitive first cell detection step~\cite{redmon2016you,he2017mask,schmidt2018,stringer2021cellpose,soelistyo_review_2023}.
However, the second linking step is still commonly solved with specialized discrete optimization methods such as integer linear programming (ILP) that allow to enforce the biological constraints on a graph data structure~\cite{jug2016moral,schiegg2013conservation,bragantini_large-scale_2023,hirsch2022weakly}. The costs for such optimization methods have to be carefully chosen, and can range from simple features such as Euclidean distance or intersection-over-union between objects~\cite{jaqaman2008,tinevez2017trackmate} to explicit motion models~\cite{ulicna_automated_2021}.
More recently, annotated tracking ground truth videos have been used to learn cell flow models with convolutional neural networks for better cost prediction~\cite{malin-mayor2023,hirsch2022weakly,hayashida_cell_2019,hayashida_mpm_2020,hayashida_consistent_2022,embedtrack2022,sugawara2022tracking}, or to directly learn object associations in a local context via graph neural networks~\cite{benhaim2022,schwartz2023} or via iterative semantic segmentation~\cite{oconnor_delta_2022}.
Furthermore in~\cite{bragantini_large-scale_2023,turetken2016network}, multiple segmentation hypotheses are considered during the discrete linking optimization problem, and a feasible unique solution is enforced with adequate additional constraints.

In this paper, we explore the potential of directly learning association costs between segmented objects using a straightforward transformer architecture.
This approach models the all-to-all interactions between objects within a short temporal window, while relying solely on shallow per-object features like position and basic shape features. Importantly, we aim to learn the correct associations even for dividing objects, in order to reduce the reliance on computationally expensive combinatorial optimizers and to potentially allow for a greedy linking post-processing only based on the learned associations.

The transformer neural network architecture \cite{vaswani_attention_2017} was originally introduced in the context of language translation, a sequence-to-sequence task, where the input is a set of tokens embedded in a continuous vector space. 
These are then fed through (self/cross) attention layers that enable the model to reason across all tokens at once. 
This particular inductive bias has produced state-of-the-art results for various predictive tasks using different elementary tokens, for example image patches as tokens for image classification~\cite{dosovitskiy_image_2020}, object detection~\cite{detr_2020} and segmentation~\cite{kirillov_segment_2023}, amino acids as tokens for protein structure prediction~\cite{alphafold2021}, and keypoints as tokens for image feature matching~\cite{lindenberger_lightglue_2023} or tasks on point clouds~\cite{transformer_pointclouds_2022}.
Notably, multiple object tracking (MOT) in the domain of natural images has recently been successfully approached with transformers in~\cite{meinhardt_trackformer_2022,transtrack2020,gtr_2022,chen2021transformer}.
Since the temporal succession of cell detections in a video can be seen as a sequence of tokens, we here investigate whether a purely  transformer-based model can be used to learn the correct associations between cell detections in a video, especially in the presence of dividing cells.

Our contributions are as follows:
\emph{i)} We provide \trackastra (\textit{Tra}cking-by-\textit{As}sociation with \textit{Tra}nsformers), which is, to the best of our knowledge,  the first plain transformer-based cell tracking method that accounts for dividing objects. Our method is simple, formulating tracking as a direct association prediction task followed by greedy linking, thereby reducing the reliance on the computationally expensive optimizer step whose hyperparameter tuning is a common practical challenge in cell tracking experiments.
\emph{ii)} We show that very simple object features such as position or basic shape features are enough for the subsequent model to learn the correct associations, which removes the need for dedicated visual feature extractors such as pretrained CNNs.
\emph{iii)} To ensure the biological consistency of the learned associations, we introduce a block\-wise \emph{parental softmax} normalization method for the association matrix, which allows for one-to-many associations, but not many-to-one.
\emph{iv)} We evaluate our approach experimentally on three datasets from different modalities and demonstrate that \trackastra outperforms even baselines that are highly tuned for specific model organisms.

\subsection{Related work}
Cell tracking in microscopy videos is closely related to the extensively studied \textit{multiple object tracking} (MOT) problem in computer vision~\cite{luo2021multiple}, for which in recent years several transformer-based approaches have been proposed~\cite{transtrack2020,meinhardt_trackformer_2022,gtr_2022,chen2021transformer}.
In contrast to MOT~\cite{leal2015motchallenge}, however, the cell tracking problem exhibits two crucial differences that motivate our work: 
First, raw image frames are typically grayscale and contain thousands of objects with very similar visual appearance that are hard to distinguish locally. 
Second, objects are allowed to split (but generally not to fuse), which introduces a more complex combinatorial solution space and precludes computationally efficient tracking approaches commonly used in MOT, such as network flows~\cite{schulter2017deep}.

\subsubsection{MOT with transformers}
In~\cite{meinhardt_trackformer_2022,transtrack2020,motr2022,transmot2023} MOT is done autoregressively end-to-end with a transformer, \ie the attention is computed between all candidate object detection tokens in frame $t$ (the keys) and the aggregated token representations of already linked tracks until frame $t-1$ (the queries).
Alternatively, the attention matrix can be computed with bi-directional temporal context between single detections only, considering all detections within a local sliding window~\cite{gtr_2022,meinhardt_novis_2023}. 
Closest to our work is the tracking-by-detection approach \textit{Global Tracking Transformers}~\cite{gtr_2022}, which, however, cannot account for cell divisions and heavily relies on appearance features of each detection that are extracted with a convolutional neural network (CNN) from the raw video frames.
\subsubsection{Learning based association prediction for cell tracking}
In microscopy image analysis, transformer-based tracking has been used in the context of single particle tracking~\cite{particle_transformer_2023} where most detections look similar like in our problem setting, however without supporting dividing objects. 
In~\cite{oconnor_delta_2022}, the association step involves densely predicting for each object its mask in the subsequent frame, which is then used to generate appropriate linking costs.  
For dividing objects, explicitly learning associations between objects has been explored with Graph Neural Networks in~\cite{schwartz2023,benhaim2022}, where object features extracted by a CNN interact via the given structure of a hypothesis graph. 
This constrains detection interactions to a certain locality, in contrast to direct all-to-all interactions in a transformer.

\section{Method}

Our proposed \trackastra method operates on raw image sequences and corresponding  detections (or segmentation masks) and uses an encoder-decoder transformer to directly predict the pairwise association matrix $A$ between all detections in a local window of consecutive time frames. 
Specifically, we construct a token for each object and timepoint within the local window and use this sequence of tokens as input to the transformer. 
The predicted associations $\hat{A}$ are then used as costs in a candidate track graph that is pruned either greedily or via  discrete optimization to obtain the final cell tracks. An overview of the full pipeline is shown in \cref{fig:method}. The following sections describe the dataset and training target construction, the transformer architecture, the loss function, the inference and final link assignment, and implementation details.

\subsection{Dataset and association matrix construction}

Let $I_1, I_2, \dots, I_T \in \R^{w \times h}$ be an image sequence that is grouped into overlapping windows $S_1, \dots, S_{T-s+1} \in \R^{s\times w\times h}$ of size $s$.
Each window $S_n$ contains a set of detections $\{d_i\}$ that each correspond to
a time point $t_i\in \N$, a center point $p_i\in \R^2$, a segmentation mask $m_i \in \{0,1\}^{w\times h}$, and other potential object features $z_i\in\R^z$ such as basic shape descriptors or mean image intensity of the instance.
The goal of the model is to predict an association probability matrix $\hat{A}=(\hat{a}_{ij})$ between all $d_i$ in the window $S_n$.
To construct the target association matrix $A=(a_{ij})$ the set of detections $\{d_i\}$ is matched to the set of ground truth objects $V = \{v_k\}$ and their ground truth associations. 
Each ground truth object again corresponds to a time point $t_k$, center point $p_k$, and a segmentation mask $m_k$ and the tracking associations can be described as a directed tree $G = (V,E)$. An edge $e_{kl}$, $k,l \in V$ exists only if $t_k + 1 = t_l$ and the objects $v_k$ and $v_l$ represent the same cell at different time points, or if $v_k$ is the mother cell of $v_l$.
As a simple matching criterion between detections $d_i$ and ground truth objects $v_k$ we use
\begin{equation}
M_{ik} = \max\left(\mathrm{IoU}(m_i, m_k), 1 - \frac{||p_i-p_k||_2}{\delta_{max}}\right) > 0.5\quad ,	
\end{equation}
where $\delta_{max}$ is a distance threshold and $\mathrm{IoU}$ denotes the intersection-over-union. The final matching is then obtained by solving a minimum cost bipartite matching problem based on the costs $M_{ik}$ between $\{d_i\}$ and $\{v_k\}$.
Finally, for all matched pairs of detections $(d_i, v_{k_i})$ and $(d_j, v_{k_j})$, we set $a_{ij} = 1$ if $v_{k_i}$ and $v_{k_j}$ are part of the same sub-lineage, \ie iff $v_{k_i} \in descendants(v_{k_j})$ or $v_{k_i} \in ancestors(v_{k_j})$, otherwise we set $a_{ij} = 0$. Note that this way, also associations across non-adjacent timepoints as well as appearing and disappearing objects are supported.

\subsection{Transformer architecture}

The input tokens $x_i \in \R^d$ are constructed by using learned Fourier spatial positional encodings $\Theta$ for the detection positions $p_i$, concatenating them with the object features $z_i$, and projecting them onto the token dimensionality $d$:
\begin{equation}
	x_i = W_{inp} [\Theta(p_i), z_i] \quad .
\end{equation}
where $z_i$ are the low-dimensional feature vector containing shallow texture and morphological features (such as mask area or mean intensity) and $W_{inp}$ is a linear projection layer mapping the concatenated tensor to $\R^d$.
The model consists of an encoder-decoder transformer architecture of $2L$ multi-head attention layers with 4 heads each~(\cf~\cref{fig:method}):
\begin{equation}
\mathcal{A}(Q, K, V) = \text{softmax}\left(\frac{QK^T}{\sqrt{d}} + M\right) V \quad, 
\end{equation}
where $Q$ are the projected attention \textit{queries}, $K$ the projected \textit{keys}, $V$ the projected \textit{values},
and $M$ is a mask disabling attention for all token pairs whose distance is larger than a user defined threshold $d_{max}$, \ie $M_{ij} = 0$ if $||p_i-p_j||_2 \leq d_{max}$ and $M_{ij} = -\infty$ otherwise. 
In every attention layer, we additionally use rotary positional embeddings (RoPE~\cite{roformer_2022})
to directly and efficiently inject relative spatial and temporal information.
The encoder $f$ transforms the input tokens using $L$ self-attention layers $\mathcal{A}_f^{\ell}(X,X,X)$
to obtain representations $Y = f(X)$.
The decoder $g$ uses $L$ cross-attention layers $\mathcal{A}_g^{\ell}(X,Y,Y)$
to obtain a second set of representations $Z = g(X, Y) $.
In between attention layers we use a simple two-layer MLP with GeLU activation, layer normalization and add residual connections following~\cite{vaswani_attention_2017}.
Finally, we apply two-layer MLPs to $Y$ and $Z$ and compute the logits $\hat{A}$ of the final association matrix as the outer product $\hat{A} = \mathrm{MLP}_Y(Y) \cdot \mathrm{MLP}_Z(Z)^T$.

\subsection{Parental softmax}
Given the predicted association logits $\hat{A}$, a simple approach to extract association probabilities $\tilde{A} \in (0,1)$ would be to apply a sigmoid to each entry of $\hat{A}$, \ie $\tilde{A} = \sigma(\hat{A})$. However, this approach does not enforce the combinatorial constraints of cell tracking, \ie the uniqueness of each object's parent while allowing for more than one child, as well as appearance and disappearance of objects.
To remedy this, we propose a logit normalization that we call \emph{parental softmax} and which ensures that the block-wise sum of all entries in the vector of possible parent associations for each $d_i$ is at most one~(\cf~\cref{fig:method}). Concretely, we define the parental softmax $\Phi(\hat{A})$ as 
\begin{equation}
	\tilde{A} = \Phi(\hat{A})_{ij} = \frac{\exp(\hat{A}_{ij})}{1 + \sum_{i'\in \mathcal{P}_{j}} \exp(\hat{A}_{i'j})}\quad ,
    \label{eq:parental_softmax}
\end{equation}
where $\mathcal{P}_j = \{d_{i'} | t_{i'} = t_j - 1,\ \forall i' \in D\}$ denotes all detections in the frame before detection $d_j$. 
Note that adding a constant to the denominator (\emph{quiet softmax}) allows for detections to not be assigned to any parent detection, accommodating for appearing and disappearing objects.
We then define the loss to be minimized during training as
\begin{align}
	\mathcal{L}(A, \hat{A}, W) &= 
	\mathcal{L}_{BCE}(A, \Phi(\hat{A}), W) + \lambda\mathcal{L}_{BCE}(A, \sigma(\hat{A}), W) \quad ,
	\label{eq:full_loss}
\end{align}
where $\mathcal{L}_{BCE}$ is the usual element-wise binary cross-entropy loss, $\lambda \in \R$ is a small fixed hyperparameter (we use $\lambda=10^{-2}$ throughout), and $W$ is a weighting factor for each matrix element.
The elementwise weighting terms $W$ are set to
\begin{equation}
 w_{ij} =
    \begin{cases}
    \begin{aligned}
      & 0 \quad & t_j - t_i > \Delta t &\qquad\qquad \textit{temporal cutoff} \\
      && \lor\ t_j - t_i < 1 &\qquad\qquad \textit{only forward links} \\	
      & 1 + \lambda_{\mathrm{div}} & \mathrm{deg}^+(v_{k_i}) = 2&\qquad\qquad \textit{dividing cells} \\
      & 1 + \lambda_{\mathrm{cont}} & \mathrm{deg}^+(v_{k_i}) = 1 &\qquad\qquad \textit{continuing tracks} \\
      & 1 & \text{otherwise} &
    \end{aligned}
    \qquad ,
    \end{cases}       
\end{equation}
where $\mathrm{deg}^+(v)$ is the out-degree of vertex $v$ in $G$. We choose $\Delta t = 2$, $\lambda_{\mathrm{div}}=10$ and $\lambda_{\mathrm{cont}}=1$ as fixed hyperparameters. This choice effectively up-weights the loss for cell divisions and continuing tracks, and removes the loss for associations that are not used during the linking step.

\subsection{Inference and linking}
Inference is done with a sliding window of size $s$ as in training. To obtain global scalar association scores $0 \leq \bar{a}_{i'j'} \leq 1$ from $\tilde{A}^{(1)}, \dots, \tilde{A}^{(T-s+1)}$, where $i'$ and $j'$ are global detection indices in a video $I_1, I_2, \dots, I_T$, we take the mean over the $s-1$ windows that include this association
\begin{equation}
	\bar{a}_{i'j'} = \frac{1}{s-1} \sum_{\{S_n | i', j' \in S_n\}} \tilde{a}_{i'j'}^{(S_n)} \quad .	
\end{equation} 
Next, we build a candidate graph $G_C=(V,E)$ with a maximum admissible Euclidean distance $dist_{max}$ between detections in adjacent time frames. For this, we use associations $\bar{a}_{i'j'}$ with $t_j' - t_i' = 1$, \ie the upper blockwise diagonal of $\bar{A}$. 
To generate a first association candidate graph we directly discard small associations with $\bar{a}_{i'j'} < \alpha$ with $\alpha = 0.05$. This candidate graph is then pruned to a solution graph $G_S=(V_S,E_S)$ with $V_S \subseteq V, E_S \subseteq E$ with one of the following linking algorithms:
\subsubsection{Greedy} We iteratively add edges and their incident nodes to $G_S$, ordered by descending edge probability, if the edge probability $\theta \geq 0.5$ and if the edge does not violate the biological constraints (\ie at most two children, and at most one parent per vertex)
\begin{equation}
	\mathrm{deg}^+(v) \leq 2 \quad \forall v \in V_S \quad,\qquad \mathrm{deg}^-(v) \leq 1 \quad \forall v \in V_S \: .
\label{eq:bio_constraints}
\end{equation}
\subsubsection{Linear assignment problem (LAP)} We use the established two-step LAP as described by Jaqaman \etal~\cite{jaqaman2008}, implemented in~\cite{fukai_laptrack_2023}. In the first step, linear chains are formed, which are connected to full cell lineages in a second step. We set a maximum linking distance adapted to the respective dataset, and use the default values for all other hyperparameters. 
\subsubsection{Integer linear program (ILP)} We solve a global ILP with all detections as graph vertices and associations $\{\bar{a}_{i'j'}\}$ with $t_{j'} - t_{i'} = 1$ as edges. The formulation enforces the biological constraints in~\cref{eq:bio_constraints}, as described in~\cite{malin-mayor2023}. We set the parameters of the ILP,~\ie the linear weights of different classes of costs, to values that balance the likelihoods of appearance, disappearance and divisions of cells.

\subsection{Implementation details}
\trackastra can be trained on a single GPU (\eg Nvidia RTX 4090) for prototypical 2D cell tracking datasets. The transformer is implemented in \textit{PyTorch}. We set window size $s=6$, embedding dimension $d=256$, number of encoder and decoder attention layers $L=6$, the maximum number of tokens per window $|D|=2048$, and batch size 8. As shallow object features $z_i$ we use the mean intensity, the object area and the inertia tensor of the object region~\cite{jahne1993spatio}. We apply the following data augmentations
jointly to all frames in a window:
flips, shifts, rotations, shear, scaling, intensity shifting and scaling, and temporal subsampling.
The augmentations are applied directly to the object features.
\trackastra association prediction scales to videos with thousands of objects, \eg, inference runs at $\sim 1$~FPS on a single GPU for 2k objects per frame.
The ILP is implemented in \textit{Motile} \cite{motile}.
\section{Experiments}

\begin{figure}[t]
    \centering
   \includegraphics[width=1.0\textwidth]{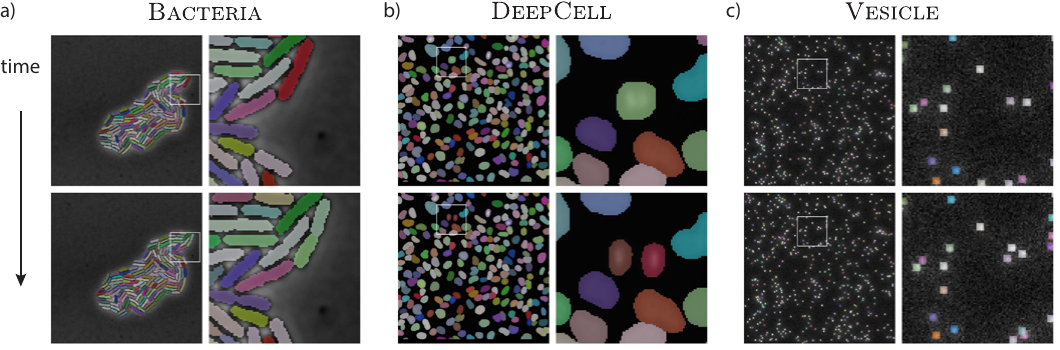}
    \caption{Cell tracking datasets evaluated in the experiments section. 
    a) \bacteria dataset from~\cite{vanvliet2018} that shows dense colonies of growing and dividing bacteria. 
    b)  \deepcell dataset of moving and dividing cells with labeled nuclei (DynamicNuclearNet from~\cite{schwartz2023}). 
    c) \vesicle dataset from the ISBI particle tracking challenge~\cite{chenouard_isbi_2014} that shows synthetically generated images of fluorescently labeled particles.
    }
    \label{fig:data}
  \end{figure}

In the following we present tracking experiments on three datasets with varying characteristics, preceded by the employed evaluation measures.

\subsection{Metrics}
We report absolute errors per video for multiple elementary error types: False positive (FP) edges, false negative (FN) edges, as well as false positive and false negative divisions.
To obtain an aggregated error measure, we further report the Acyclic Oriented Graphs Matching (AOGM) measure~\cite{matula_cell_2015}, which accounts for the number of operations to transform a predicted graph into a reference ground truth (GT) graph. To start, predicted detections are matched to reference detections if they cover more than half of a reference detection's area.
After that, the following operations are performed in the given order to transform the predicted graph into the reference graph:
\begin{enumerate}
	\item $NS$: (\textit{node split}) Split predicted node that false merged multiple reference nodes, and delete the incident edges of the original node (does not occur when linking GT detections).
	\item $FN$: Add FN node.
	\item $FP$: Delete FP node and its incident edges (does not occur when linking GT detections).
	\item $ED$: Delete FP edge.
	\item $EA$: Add FN edge.
	\item $EC$: Change semantics of predicted edge (linear chain link \vs division link). 
\end{enumerate}
$\mathrm{AOGM}$ is then defined as the weighted sum of all operations (following \cite{ulman_objective_2017})
\begin{equation}
	\mathrm{AOGM} = 5\cdot|NS| + 10\cdot|FN| + 1\cdot|FP| + 1\cdot|ED| + 1.5\cdot|EA| + 1\cdot|EC|\: ,
\end{equation}
which represents the quality of a cell tracking solution for a given video in terms of errors to tolerate, in contrast to the usual dominant fraction of simple correctly assigned links. Note that cell divisions are counted implicitly in AOGM.
We also report $\mathrm{TRA}$, which is the normalized version of AOGM commonly used in cell tracking~\cite{ulman_objective_2017}
\begin{equation}
	\mathrm{TRA} = 1 - \frac{\min(\mathrm{AOGM},\mathrm{AOGM_0})}{\mathrm{AOGM_0}}, \qquad 0 \leq \mathrm{TRA} \leq 1 \quad ,
\end{equation}
where $\mathrm{AOGM_0}$ is the AOGM of an empty predicted graph.
Finally, when using predicted segmentations, we also report AOGM$+$, which removes the constant penalty incurred by missing detections.

\begin{figure}[t]
  \centering
\includegraphics[width=1.0\textwidth]{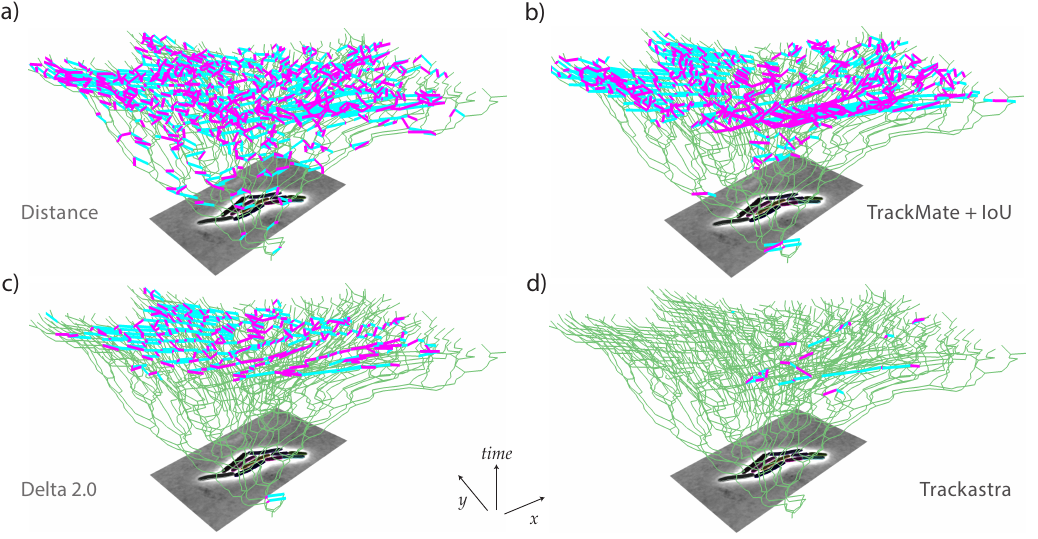}
  \caption{Error trees on a challenging \bacteria test video. Time on the vertical axis, edges colored as true positive (green), false positive (magenta) and false negative (cyan).
}
  \label{fig:errors}
\end{figure}

\subsection{Bacteria colony tracking (\bacteria)}

\label{sec:bacteria}
Here we use one of the largest public bacteria tracking datasets~\cite{vanvliet2018}. 
This dataset (denoted \bacteria) consists of 39 videos of six different types of bacteria colonies, containing roughly 100k cells with full segmentation and tracking annotations and showing roughly 9k divisions. 
This dataset is challenging, as objects are densely packed and the colonies grow and divide, leading to large displacements between frames (\cf~\cref{fig:data}a).
We split it into 31 training, two validation and six test videos, and compare our approach on ground truth detections against various baselines, including \textsc{Delta 2.0}~\cite{oconnor_delta_2022} that was explicitly created for bacterial colony tracking (\cref{tab:bacteria}). 
\begin{table}[t]
	\setlength{\tabcolsep}{3pt}
  \centering
  \caption{Tracking results on \bacteria (using ground truth detections). The test set consists of six videos with a mean of 2912 edges and 236 divisions. $\downarrow$--metrics report mean absolute errors per video. We show results for three runs per model. The last two rows show results of the general model (\cf~\cref{sec:multidomain}).
  }
  \begin{adjustbox}{max width=\textwidth}
\begin{tabular}{ll S[table-format=1.3] S[table-format=-4] S[table-format=4] S[table-format=3] S[table-format=3] S[table-format=3] S[table-format=3]}
\toprule
                               Method & Linking &{TRA$\:\uparrow$}&{AOGM$\:\downarrow$}&{FP edges$\:\downarrow$}&{FN edges$\:\downarrow$}&{FP divs$\:\downarrow$}&{FN divs$\:\downarrow$} \\
\midrule
                          Distance & greedy 		& 0.944 			&  1511 &       447 &       451 &      298 &      194 \\
                          Distance & ILP 		& 0.962 			&  1065 &       303 &       303 &      165 &      193 \\
TrackMate (overlap)~\cite{tinevez2017trackmate} & LAP  & 0.957 			&   872 &       256 &       292 &       77 &       50 \\
            \textsc{Delta} 2.0~\cite{oconnor_delta_2022} & greedy 		& 0.996 			&   118 &        40 &        43 &       12 &        4 \\
       \trackastra (points only) & ILP 			& 0.995 			&   136 &        39 &        39 &       20 &       29 \\
				\trackastra & greedy 			& \bfseries 0.999&    36 &        10 &        12 &        5 &        \bfseries 3 \\
					\trackastra & ILP 			& \bfseries 0.999&    23& 7 &		8&\bfseries 2& \bfseries 3 \\
					
\midrule
\trackastra-general & greedy 			&  		\bfseries 0.999 &   29 &     8 &     9 &    4 &    \bfseries 3 \\
\trackastra-general & ILP 			& 	 	\bfseries 0.999 &   \bfseries 19 &   \bfseries  6 &  \bfseries   7 &   \bfseries 2 &  \bfseries  3 \\

\bottomrule
\end{tabular}
   \end{adjustbox}
  \label{tab:bacteria}
\end{table}
As expected, simple Euclidean-distance-based tracking achieves poor scores, even if a powerful integer linear programming (ILP) linker is used, since the frame rate in this dataset is low and the objects move notably from frame to frame (\cf~\cref{fig:data}a, \cref{fig:errors}a).
A standard procedure for bacteria tracking using the accordingly configured and widely used cell tracking tool TrackMate~\cite{tinevez2017trackmate} also leads to poor results, since the movements are often too large even for a generously configured intersection-over-union-based tracker (\cf~\cref{fig:errors}b).
The state-of-the-art deep-learning-based bacteria tracking algorithm \textsc{Delta 2.0}~\cite{oconnor_delta_2022}, trained on the same dataset as \trackastra, reduces the total AOGM notably, as shown in~\cref{fig:errors}c.
Surprisingly, training \trackastra using only center point coordinates as features (\ie without any appearance information) performs only slightly worse than \textsc{Delta 2.0}, which in turn does make use of both the grayscale images and the segmentation masks. This highlights the strong cues that object motion alone can provide for biological datasets.  
When training \trackastra with shallow object features, already using greedy linking based on the predicted associations reduces the total errors (AOGM) by $\sim 70\%$ compared to the state of the art (\cref{tab:bacteria}, \cref{fig:errors}d, Vid.~S1). When adding an ILP linker, we are further reducing the total error to an almost perfect tracking result ($\mathrm{AOGM}=23$ \vs $118$ for \textsc{Delta 2.0}).

\subsection{Nuclei tracking (\deepcell)}
\label{sec:deepcell}

Next, we evaluate \trackastra on DynamicNuclearNet~\cite{schwartz2023}, the largest publicly available dataset for cell nuclei tracking. 
This dataset (here denoted \deepcell) contains 130 videos of fluorescently labeled nuclei from five different cell types, containing roughly 600k cells with full segmentation and tracking annotations and showing roughly 2k divisions.
Due to the comparatively lower rate of cell divisions, we choose to reduce the impact of divisions in the loss reweighting $W$ by setting $\lambda_{\mathrm{div}} = 2$ for this dataset.
We adhere to the training-validation-test split as defined in~\cite{schwartz2023}.
We train a \trackastra model based on ground truth segmentations, as well as a model based on the segmentations from the Caliban pipeline~\cite{schwartz2023} (\cf~\cref{tab:deepcell}).
We compare our approach to two recent methods that learn linking costs explicitly~\cite{schwartz2023,benhaim2022}, and take the scores reported in~\cite{schwartz2023}.

Using ground truth segmentations \trackastra generally outperforms Caliban~\cite{schwartz2023}, the state\hyp{}of\hyp{}the\hyp{}art model that is specifically tuned for \deepcell (\cf~\cref{tab:deepcell}). Specifically the total number of errors (AOGM) is more than halved when using an ILP ($\mathrm{AOGM}=7.9$ \vs $18.1$), and shows already a notable reduction when using a simple greedy linker ($\mathrm{AOGM}=11.9$ \vs $18.1$, \cf~Vid.~S2). However, Caliban slightly outperforms \trackastra in terms of Division F1, which might be due to explicit modeling of division events in Caliban.
We additionally show in \cref{tab:deepcell} results of CellTrackerGNN~\cite{benhaim2022} as reported in \cite{schwartz2023}, which are worse than both Caliban and \trackastra. Note that the available CellTrackerGNN models have been trained on datasets from the Cell Tracking Challenge~\cite{maska_cell_2023}, which might explain their worse performance on \deepcell.

\begin{table}[t]
\setlength{\tabcolsep}{3pt}
\centering
    \caption{Tracking results on \deepcell. The test set consists of 12 videos with a mean of 4018 edges and 15 divisions. Scores for Baxter, Caliban and CellTrackerGNN as evaluated in~\cite{schwartz2023}. Metrics represent mean per test video.
    AA: Association accuracy~\cite{hayashida_mpm_2020}.
    AOGM$+$ removes the constant penalty due to missing detections (498.6).}
 \label{tab:deepcell}

\begin{adjustbox}{max width=\textwidth}
\begin{tabular}{ll H @{\hskip 20pt} S[table-format=3.1] ccc wc{8pt}  H H S[table-format=3.1] S[table-format=3.1] ccc}

\toprule
&&& \multicolumn{4}{c}{Ground truth objects} &&& \multicolumn{6}{c}{Caliban segmentations~\cite{schwartz2023}}\\
\cmidrule(lr){4-7} \cmidrule(lr){10-15}
Costs						& Linking 		&{AOGM total$\:\downarrow$} &{AOGM$\:\downarrow$}	&TRA$\:\uparrow$&Div F1$\:\uparrow$	& AA$\:\uparrow$&&{AOGM total$\:\downarrow$}&{AOGM-A$\:\downarrow$}&{AOGM$+\downarrow$}&{AOGM$\:\downarrow$}&{TRA$\:\uparrow$}&{Div F1$\:\uparrow$}	&{AA$\:\uparrow$}  \\
\midrule
Baxter~\cite{magnusson2014}	& greedy			& {-}						& {-}						& 0.997 		& 0.72				& \textbf{1.00}	&&& {-}					&{-}&{-}					& 0.987			& 0.60				& \textbf{0.98} \\
CellTrackerGNN~\cite{benhaim2022}& greedy	& 1540						& 128.3 						& 0.999			& 0.18				& 0.93 			&& 8435				& 245.8 & 204.3 & 702.9 				& 0.988			& 0.13 				& 0.89			\\
Caliban~\cite{schwartz2023}	& LAP			& 217						&	18.1						& \textbf{1.000}& \textbf{0.97}		& 0.99 			&&& 154.2 			& 		75.4 	& 574.0			& \bfseries 	0.991			& \textbf{0.92}		& 0.97			\\

\trackastra				& greedy						& 143				& 11.9				& \textbf{1.000}	& 0.90			& \textbf{1.00} 			&& 7101				& 148.2 & 93.2 & 591.8					& \textbf{0.991}& 0.71				& 0.96			\\
\trackastra				& ILP				& \bfseries 95				&  7.9				& \textbf{1.000}	& 0.94			& \textbf{1.00} 			&& \textbf{6785}	& 145.4 & 66.8 & 565.4					& \textbf{0.991}& 0.79				& \textbf{0.98} \\

\midrule
\trackastra-general & greedy				&	& 7.4 			& \bfseries 1.000 &        0.96 & \bfseries 1.00 &     		&& 146.0 &   			67.8 &		566.4 			& \bfseries 0.991 & 0.59 & \bfseries 0.98 \\
\trackastra-general & ILP				&   & \bfseries 5.8 & \bfseries 1.000 &        0.94 & \bfseries  1.00  &    	&& \bfseries 145.0 &   \bfseries 66.6   &  	\bfseries 565.2	& \bfseries 0.991 & 0.65 & \bfseries 0.98\\

\bottomrule
\end{tabular}
\end{adjustbox}
\end{table}

 On predicted segmentations (using the provided masks from~\cite{schwartz2023}), our best in-domain model again outperforms Caliban in terms of total errors after removing the constant penalty for missing detections ($\mathrm{AOGM+}=66.8$ \vs $75.4$, \cf~\cref{tab:deepcell}).
However, \trackastra is again not able to reach the Division F1 of Caliban. As before, CellTrackerGNN models have been trained on datasets from the Cell Tracking Challenge, and are therefore not performing well on \deepcell. 
\begin{table}[t]
	\setlength{\tabcolsep}{3pt}
  \centering
  \caption{Out-of-domain results on \hela.
  Specialized models trained only on \deepcell (using ground truth detections), \trackastra-general trained on a diverse dataset. The test set contains two videos with a mean of 16835 edges and 151 divisions.}
  \begin{adjustbox}{max width=\textwidth}
\begin{tabular}{l @{\hskip 20pt} l S[table-format=1.3] S[table-format=3] H S[table-format=2] S[table-format=3] S[table-format=2] S[table-format=3] }
\toprule
            {Method} &  Linking & {TRA$\:\uparrow$} &  {AOGM$\:\downarrow$} &  {switches$\:\downarrow$} &  {FP edges$\:\downarrow$} &  {FN edges$\:\downarrow$} &  {FP divs$\:\downarrow$} &  {FN divs$\:\downarrow$} \\
\midrule
Caliban~\cite{schwartz2023} & LAP	& 0.994 				&   931 				&  0  &	 \bfseries 36 	&  275 			&       61 			&      151 \\
\trackastra  & greedy 				& 0.999 				&   254 				&  73 &  52 			&  	94 			&       24 			&       41 \\
\trackastra  & ILP 					& 0.999 				&   190 				&  59 &  72 			&  	48 			&       35 			&       17 \\
\midrule
\trackastra-general & greedy & 0.999 & 145 &   NA &   48 &      25 &     22 &      4\\
\trackastra-general & ilp  &   \bfseries 1.000 & \bfseries 96 &     NA& 		47 &   \bfseries   24 &   \bfseries  16 &  \bfseries    4\\
\bottomrule
\end{tabular}
 	\end{adjustbox}
 	\label{tab:hela}
\end{table}

To probe the out-of-domain capabilities of \trackastra, we take a model trained on \deepcell and apply it to a similar dataset of fluorescently tagged nuclei of HeLa cells from the Cell Tracking Challenge (denoted \hela), refer to~\cref{tab:hela}. 
Interestingly, \trackastra significantly outperforms Caliban~($\mathrm{AOGM} = 190$ \vs $931$), potentially because of its use of only shallow input features (such as positions and basic shape descriptors), which likely helps the model prevent overfitting.
Notably, the general model (\trackastra-general, \cf \cref{sec:multidomain}) trained on a diverse dataset performs substantially better than the specialized models.

\subsection{ISBI particle tracking challenge}
\label{sec:particle}

\begin{table}[t]
	\setlength{\tabcolsep}{3pt}
    \centering
    
    \caption{Tracking results on the \vesicle dataset from the ISBI particle tracking challenge (using GT detections). For a definition of the metrics see~\cite{chenouard_isbi_2014}. 
    }
    \scriptsize
    \begin{tabular}{ll S[table-format=1.3] S[table-format=1.3] S[table-format=1.3] S[table-format=1.3] S[table-format=1.3] S[table-format=1.3] S[table-format=1.3] S[table-format=1.3] S[table-format=1.3]}
        
\toprule
&& \multicolumn{3}{c}{Vesicles (low)} & \multicolumn{3}{c}{Vesicles (mid)} & \multicolumn{3}{c}{Vesicles (high)}\\
\cmidrule(lr){3-5}  \cmidrule(lr){6-8} \cmidrule(lr){9-11} 
    Density & Method &  $\alpha\uparrow$ &  $\beta\uparrow$ &  $JSC_\theta\uparrow$ & $\alpha\uparrow$ &  $\beta\uparrow$ &  $JSC_\theta\uparrow$  & $\alpha\uparrow$ &  $\beta\uparrow$ &  $JSC_\theta\uparrow$ \\
\midrule
     & LAP~\cite{jaqaman2008,trackmate2022}   & \textbf{0.953} & \textbf{0.947} & \textbf{0.979}  & 0.753 & 0.703 & 0.704 & 0.568 & 0.490 & 0.515 \\
low  & KF~\cite{kalman1960}    & 0.937 & 0.924 & 0.959 & 0.673 & 0.609 & 0.787  & 0.477 & 0.389 & 0.643  \\
     & MoTT~\cite{particle_transformer_2023} & 0.926 & 0.891 & 0.925 & \textbf{0.800} & 0.733 & 0.874  & 0.652 & 0.544 & 0.748\\
     & \trackastra & 0.945 & 0.936 &  0.972  & 0.798 & \textbf{0.757} & \textbf{0.895} & \textbf{0.672} & \textbf{0.602} &  \textbf{0.806}\\
\bottomrule
\end{tabular}
\label{tab:particle}
\end{table}

Here we assess how well \trackastra performs for single molecule tracking, a domain closely related to cell tracking where, however, objects do not divide. 
Specifically, we use the challenging \vesicle dataset from the ISBI particle tracking challenge~\cite{chenouard_isbi_2014} (\cf~\cref{fig:data}c).  
This gives us the chance to compare \trackastra against a state-of-the-art transformer-based tracking approach, MoTT~\cite{particle_transformer_2023}, as well as classical methods such as the Kalman filter (KF)~\cite{kalman1960} and the linear assignment problem (LAP)~\cite{jaqaman2008,trackmate2022} (\cf~\cref{tab:particle}). 
Interestingly, even in this context \trackastra outperforms MoTT for the case of high vesicle density when using ground truth detections in terms of the association accuracy ($\alpha, \beta$) as well as Jaccard similarity coefficient ($JSC_\theta$), which is a measure of the overlap between the predicted and ground truth tracks~(\cref{tab:particle}).
This highlights the general applicability of our approach, as no adjustments to the model or the training procedure were required.

\subsection{Multi-domain general \trackastra model}
\label{sec:multidomain}
We train a \trackastra model on a diverse dataset, including all data used in the experiments described before (\cref{sec:bacteria,sec:deepcell,sec:particle}), as well as additional data from the Cell Tracking Challenge \cite{maska_cell_2023} and from~\cite{ker_phase_2018,obiwan,zargari2023deepsea,funke2018benchmark}. We evaluate this model (denoted \trackastra-general) on the \bacteria and \deepcell test sets.
Notably, on both \bacteria and \deepcell, the general model outperforms the specialized single-domain \trackastra models (\cf~\cref{tab:bacteria,tab:deepcell}).
Furthermore, when applying the general model to the out-of-domain \hela dataset (which was not part of any training set), it outperforms the specialized \trackastra model trained on \deepcell, demonstrating the importance of a large, diverse training set for out-of-domain tracking performance (\cf~\cref{tab:hela}).

Finally, we also submit a single \trackastra-general model to the cell linking benchmark of the Cell Tracking Challenge (7th edition, 2024). Our model performs best overall according to structural as well as biological metrics across 13 diverse datasets, as well as best individually for roughly half of the datasets \cite{ctc2024}.

\subsection{Ablations}

\paragraph{Parental softmax:}
Here we explore whether the proposed parental softmax improves performance for dividing objects. 
Specifically, we compare in~\cref{tab:softmax} results on \bacteria with and without using parental softmax, \ie when using only sigmoid normalization by setting $\lambda=1$ and removing the first parental loss component in \cref{eq:full_loss}.
We find that the parental softmax reduces the number of errors (AOGM) by $\sim 20\%$ for both a greedy and an ILP linker on \bacteria, demonstrating the positive impact of the parental softmax on the tracking performance for dividing objects.
\begin{table}[t]
\scriptsize
\setlength{\tabcolsep}{3pt}
    \centering
    \caption{Ablation for parental softmax on \bacteria (using GT detections). Test set as in \cref{tab:bacteria}. We report results for two runs per model.}
\begin{tabular}{lc @{\hskip 10pt}|@{\hskip 10pt}S[table-format=2.1] H S[table-format=1.1] S[table-format=1.1] S[table-format=1.1] S[table-format=1.1]}
\toprule
Linking &Parental softmax &  {AOGM$\:\downarrow$} &  {Switches$\:\downarrow$} &  {FP edges$\:\downarrow$} &  {FN edges$\:\downarrow$} &  {FP divs$\:\downarrow$} &  {FN divs$\:\downarrow$} \\
\midrule
 greedy & \ding{55}  &  28.4 &      10.7 &      7.6 &      9.5 &     4.6 &     3.4 \\
 greedy & \ding{51}  &  23.0 &      10.0 &      5.6 &      8.4 &     3.8 &     \bfseries 1.8 \\
    ILP & \ding{55}  &  18.8 &      5.5 &      5.4 &      6.2 &     2.4 &     3.1 \\
    ILP & \ding{51}  &  \bfseries 14.8 &  \bfseries 5.1 &   \bfseries 4.2 &      \bfseries 5.2 &     \bfseries 1.8 &     \bfseries 1.8 \\
\bottomrule
\end{tabular}
    \label{tab:softmax}
\end{table}
\begin{figure}[b]
\centering
\includegraphics[width=\textwidth]{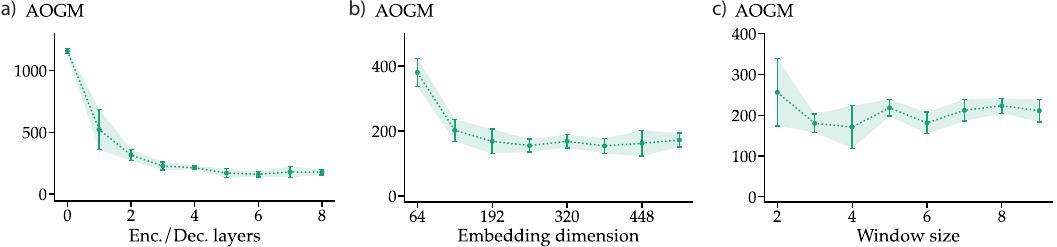}
\caption{Ablations on \bacteria (using ground truth detections) using only center points as features, and a LAP linker. Lower is better. We show results for three runs per model.
}
\label{fig:ablations}
\end{figure}
\paragraph{Transformer size:} Here we vary the number $L$ of attention layers in both the encoder and decoder, as well as the embedding dimension $d$, using slightly smaller \trackastra models for computational efficiency (\cf~\cref{fig:ablations}a,b). 
As expected, using no attention layers ($L=0$), \ie directly predicting the association matrix with the two projection heads leads to poor performance, confirming the importance of attention across all detections. Furthermore, using more than $L=6$ layers does not lead to a notable improvement.
\paragraph{Window size:} We run experiments with different window sizes $s$, \ie different temporal context available to the model (\cf~\cref{fig:ablations}c). When using only $s=2$ we find that the performance substantially deteriorates, whereas window sizes $s \in (3,6)$ all lead to good results, demonstrating that a relatively small temporal context is already enough for decent tracking results.
\begin{table}[t]
\setlength{\tabcolsep}{3pt}
    \centering
    \caption{Ablations on \bacteria (using GT detections). Test set as in \cref{tab:bacteria}. We show results for three runs per model.}
    \begin{adjustbox}{max width=\textwidth}
\begin{tabular}{wc{35pt}wc{35pt}wc{35pt} |@{\hskip 10pt} c S[table-format=3.1]  S[table-format=2] S[table-format=2] S[table-format=2] S[table-format=2]}
\toprule
\makecell{Rel.~pos.\\ encoding} & \makecell{Object\\ features} &        ILP &   {TRA$\:\uparrow$}&{AOGM$\:\downarrow$}&{FP edges$\:\downarrow$}&{FN edges$\:\downarrow$}&{FP divs$\:\downarrow$}&{FN divs$\:\downarrow$} \\
\midrule
              &                &            & 0.989 &   314 &        72 &        88 &       35 &       44 \\
              &                & \checkmark & 0.992 &   235 &        72 &        73 &       28 &       42 \\
   \checkmark &                &            & 0.994 &   164 &        42 &        47 &       24 &       30 \\
   \checkmark &                & \checkmark & 0.995 &   136 &        39 &        39 &       20 &       29 \\
   \checkmark &     \checkmark &            & \bfseries 0.999 &    36 &        10 &        12 &        5 &        \bfseries 3 \\
   \checkmark &     \checkmark & \checkmark & \bfseries 0.999 &    \bfseries 23 &         \bfseries 7 &         \bfseries 8 &        \bfseries 2 &     \bfseries 3 \\
\bottomrule
\end{tabular}
     \end{adjustbox}
    \label{tab:ablation}
\end{table}
\paragraph{Other components:} Finally, we ablate basic components of \trackastra in~\cref{tab:ablation}. Removing positional encodings as well as the basic shape descriptors from \trackastra increases errors notably, while the decrease in performance when replacing the tailored ILP optimizer with a simple greedy linker is less pronounced.

\section{Discussion}

We presented \trackastra, a robust method to track dividing objects such as cells using a powerful and scalable transformer architecture.
\trackastra uses only object positions and shallow object features as input, making it readily applicable to new scenarios whenever a domain specific detection method is already available.
We demonstrate that \trackastra performs well across different imaging modalities and biological model systems that share the particularity of dividing, yet non-fusing objects.
Additionally, we show that a single model trained on a large cross-modality dataset is able to generalize well to datasets from various different domains, which is crucial for practitioners.
Finally, \trackastra has the potential for tracking other types of dividing objects, such as icebergs in satellite imagery.
An important limitation of the presented approach is that it currently does not correct faulty detection inputs, but we anticipate that training a detection model and \trackastra end-to-end will address this issue.
Furthermore, we currently do not make use of the pairwise association predictions from non-adjacent time frames, which we expect to be beneficial for circumventing faulty detections, for example for track gap closing in case of missing detections.
Moreover, while the presented results are limited to 2D datasets, \trackastra is expected to scale well to 3D datasets since the architecture does not require to process dense images.
Overall our work demonstrates the potential that pure transformer-based architectures can hold for the field of cell tracking.

\section*{Acknowledgements}

We thank Arlo Sheridan, Talmo Pereira, Uwe Schmidt and Albert Dominguez for helpful discussions, Morgan Schwartz for releasing the code for the Caliban benchmarks, Simon van Vliet and Johannes Seiffarth for creating large-scale annotated datasets, and the EPFL School of Life Sciences ELISIR program and CARIGEST SA for their generous funding. Additionally we would like to acknowledge the Janelia Trackathon 2023 and the resulting metrics library \textit{traccuracy}. 

\bibliographystyle{splncs04}

\end{document}